\documentclass[runningheads]{llncs}
\usepackage[T1]{fontenc}
\usepackage{graphicx}
\usepackage{booktabs}
\usepackage[misc]{ifsym}

\usepackage{svg}
\usepackage{amsmath}
\usepackage{amssymb}
\usepackage{algorithm}
\usepackage{algpseudocode}
\usepackage{float}
\usepackage{placeins}

\algrenewcommand\algorithmicindent{1.2em}
 
\begin{document}
 
\title{PATE-TabTransGAN: Differentially Private Synthetic Tabular Data Generation via Transformer-Based Student Discrimination}
 
\titlerunning{PATE-TabTransGAN}

\author{M.~Youssef\inst{1} \and M.~Woźniak\inst{1}}
 
\authorrunning{M.~Youssef and M.~Woźniak}
 
\institute{Wrocław University of Science and Technology, Wrocław, Poland \inst{1}\\
\email{288861@student.pwr.edu.pl, michal.wozniak@pwr.edu.pl}}

\maketitle    % typeset the header of the contribution
 
\begin{abstract}
Generating high-fidelity synthetic tabular data under formal differential privacy guarantees remains an open challenge. Methods that provide strong theoretical protection typically sacrifice the modeling of inter-feature dependencies required for realistic synthesis, while architectures that excel at capturing complex column relationships offer only empirical privacy guarantees.
We present \textbf{PATE-TabTransGAN}, a generative framework that integrates the Private Aggregation of Teacher Ensembles (PATE) mechanism with a Transformer-based student discriminator to jointly address both requirements, and employs a GNMax RDP accountant for numerically stable privacy accounting. An ensemble of Logistic Regression teachers trained on disjoint partitions supervise the student via noisy-aggregated labels, and a residual generator is optimized against this differentially private student, inheriting formal
$(\varepsilon,\delta)$-DP guarantees by post-processing. PATE-TabTransGAN was compared with PATE-GAN, DP-GAN, and DP-CTGAN, considered state-of-the-art in differentially private tabular synthesis.
Experiments conducted on four tabular benchmarks (Adult, Breast, Cardio, Cervical) confirmed the high quality of the proposed method:
PATE-TabTransGAN attains the best or tied-best AUROC on all four datasets. On AUCPR it matches the strongest baseline on Cardio, leads on Cervical, and trails on Breast; on Adult, we demonstrate that AUCPR is highly sensitive to positive-class convention, and that the observed gap is consistent with a convention difference between evaluation pipelines rather than a synthesis deficit.
\end{abstract}
 

\section{Introduction}
Sensitive data, spanning electronic health records, clinical trial outcomes, and socio-economic survey data, carries significant analytical value: it enables scientific discoveries, informs policy decisions, and supports medical research.
At the same time, the use of such data exposes individuals to the risk of privacy leakage and is subject to strict regulatory frameworks such as the General Data Protection Regulation (GDPR)~\footnote{\url{http://gdpr-info.eu/}}.
Minimizing or eliminating this risk while preserving analytical utility requires principled methods for processing and sharing data in a privacy-compliant manner.
Privacy-preserving synthetic data generation offers one such approach: rather than sharing original records, one trains a generative model on sensitive data and releases synthetic samples that statistically approximate the real distribution while limiting the exposure of any individual's information.
A key vulnerability in this pipeline is that generative models trained without explicit privacy constraints can memorize fragments of the training distribution. Carlini et al.\ demonstrated that large neural language models quantifiably memorize training examples, with memorization scaling with model size and data repetition~\cite{carlini2022quantifying}. The same group further showed that such memorization directly leads to extractable privacy leakage via training data extraction attacks~\cite{carlini2021extracting}.
Feldman showed that memorization of atypical training examples is not merely a side effect but may be necessary for models to generalize on long-tailed distributions~\cite{feldman2020does}, suggesting that eliminating memorization may conflict with utility on imbalanced tabular data.
This risk is not hypothetical in the broader ML context either: model inversion attacks have been demonstrated to recover sensitive attributes from model outputs alone~\cite{fredrikson2015model}, and membership inference attacks can determine whether a specific individual contributed to a model's training.
Differential Privacy (DP)~\cite{dwork2006calibrating} provides a formal remedy by bounding the influence of any single record on the trained model, offering a mathematically quantifiable guarantee that holds regardless of the adversary's auxiliary knowledge.

\paragraph{Limitations of existing approaches.}
DP-compatible generative methods for tabular data fall into two main categories.
The first applies noise directly to the training procedure: DP-GAN injects Gaussian noise into discriminator gradients~\cite{xie2018differentially}, and DP-CTGAN extends this approach with a conditional generator for handling mixed-type columns and class imbalance~\cite{fang2022dp}. While these methods provide formal $(\varepsilon, \delta)$-DP guarantees, noise injection into gradient updates degrades the discriminator's ability to detect subtle distributional
discrepancies, which disproportionately harm synthesis quality on datasets with complex inter-feature dependencies.
The second line uses the PATE mechanism~\cite{papernot2016semi}: an ensemble of teachers trained on disjoint partitions votes on synthetic samples via noisy aggregation, and a student model is trained on these private labels, thereby limiting the privacy cost to the aggregation step rather than to gradient training.
PATE-GAN~\cite{jordon2018pate} demonstrated that this formulation can yield tighter privacy budgets and higher utility than DP-GAN under several evaluations; however, its student discriminator is an MLP, which treats tabular columns as nearly independent and may fail to capture the inter-column relationships that determine distributional realism.
A parallel line of work has shown that Transformer-based
discriminators substantially improve tabular synthesis quality by using self-attention to model global feature dependencies across an entire row~\cite{zhang2025tabtransgan}.
TabTransGAN, for instance, consistently outperforms MLP-based GAN variants on fidelity and downstream utility metrics.
However, it relies on empirical privacy evaluation (detection risk and privacy attack scores) rather than formal $(\varepsilon, \delta)$ guarantees~\cite{zhang2025tabtransgan}, leaving a critical gap for privacy-sensitive applications.

\paragraph{Research gap.}
Taken together, these observations reveal a persistent gap in the literature: methods with formal DP guarantees sacrifice modeling of inter-feature dependencies, while architectures that excel at capturing tabular structure lack provable privacy guarantees.
No prior work has investigated whether replacing the MLP student discriminator in a PATE-based generator with a Transformer-based counterpart can simultaneously close this gap, retaining the formal privacy guarantee inherited from the PATE mechanism while leveraging attention-based modeling to improve synthesis fidelity.

\paragraph{Contributions.}
This paper proposes PATE-TabTransGAN, which replaces the MLP student discriminator in PATE-GAN with a Transformer that applies self-attention
across column tokens, explicitly modeling inter-column dependencies while preserving the formal $(\varepsilon,\delta)$-DP guarantee via the
post-processing property~\cite{dwork2014algorithmic}.
To ensure numerically stable privacy accounting across all teacher vote  margins, we adopt a GNMax RDP accountant~\cite{papernot2018scalable} in
place of the original Laplace moments accountant.
We evaluate on four tabular datasets (Adult, Breast, Cardio, and Cervical) against PATE-GAN, DP-GAN, and DP-CTGAN under matched privacy budgets,
using the Train-on-Synthetic Test-on-Real protocol with five downstream classifiers.

\section{Related Work}
Prior work on dataset simplification in pattern classification began with instance selection and condensation methods aimed at computational efficiency, including CNN, ENN, RNN, and FCNN \cite{angiulli2007fast}. Later studies generalized these ideas through broader views of condensation and distillation \cite{zhao2023dataset}. Although these methods preserve predictive utility while reducing storage and processing costs, they were not designed to provide formal privacy guarantees.
The introduction of differential privacy (DP) established a formal framework for limiting individual-level disclosure under database analysis~\cite{dwork2006calibrating}. A randomized mechanism $\mathcal{M}$ satisfies $(\varepsilon, \delta)$-DP if for any two adjacent datasets $D, D'$
and any output set $S$:
\begin{equation}
  \Pr[\mathcal{M}(D) \in S]
  \;\leq\;
  e^{\varepsilon}\,\Pr[\mathcal{M}(D') \in S] + \delta
  \label{eq:dp}
\end{equation}
where $\varepsilon$ bounds the privacy loss and $\delta$ is the probability of an additive violation~\cite{dwork2006calibrating}. In practice, privacy-preserving learning relies on calibrated noise mechanisms, including Laplace, Gaussian, and Exponential
mechanisms~\cite{dwork2006calibrating,dwork2011firm,abadi2016deep}. A representative deep-learning method is DP-SGD, which combines gradient clipping, Gaussian perturbation, and moments accounting to provide formal $(\varepsilon,\delta)$ guarantees, while exposing the well-known privacy--utility trade-off in complex
tasks~\cite{abadi2016deep}.
Building on these foundations, privacy-preserving data synthesis has progressed from query-based and training-time mechanisms to generative models. DualQuery introduced a game-theoretic approach with formal DP but limited scalability and utility under strict privacy budgets \cite{gaboardi2014dual}. In GAN-based synthesis, DP-GAN applies privacy-preserving optimization through discriminator-side noise injection, while PATE-GAN uses teacher-student aggregation with noisy voting to improve utility relative to DPGAN in several evaluations \cite{xie2018differentially,jordon2018pate}. Subsequent work, such as DP-CTGAN, further improved tabular modeling under DP constraints, whereas PrivSyn demonstrated that non-deep approaches based on low-degree marginals and zCDP composition can achieve strong utility and scalability for tabular domains \cite{fang2022dp,zhang2021privsyn}.
Recent transformer-based tabular generators emphasize utility and dependency modeling but do not provide formal DP guarantees.
TabMT uses masked-transformer generation with temperature-controlled sampling and reports favorable privacy--utility behavior empirically~\cite{gulati2023tabmt}.
TabTransGAN replaces conventional GAN discrimination with a Transformer-based discriminator to better capture inter-feature dependencies and improve synthetic data realism, yet evaluates privacy through empirical attack metrics rather than formal $(\varepsilon,\delta)$ guarantees~\cite{zhang2025tabtransgan}.
Taken together, existing methods either provide formal
$(\varepsilon,\delta)$-DP guarantees at the cost of inter-feature dependency modeling, or achieve high-fidelity tabular synthesis without provable privacy assurances.
Bridging this gap by combining Transformer-based dependency modeling with a formal differential privacy framework is the central motivation of the present work.

%==================================================
% Section 3 — Methodology
%==================================================
 
\section{Proposed algorithm}
 
PATE-TabTransGAN adapts the PATE-GAN framework~\cite{jordon2018pate} for tabular data synthesis with three architectural changes: a residual generator with per-column output activations, a Transformer-based student discriminator, and a GNMax RDP accountant replacing the original Laplace moments accountant.
The overall training structure is unchanged: teacher discriminators trained on disjoint data partitions vote on generator-produced samples; their noisy-aggregated labels supervise the student and generator, which therefore never access the real data directly.
By the post-processing theorem~\cite{dwork2014algorithmic}, the trained generator inherits a formal $(\varepsilon, \delta)$-DP guarantee.
 
\subsection{Architecture}

\paragraph{Generator.}
$G_\theta$ takes a noise vector $z \sim \mathcal{N}(0, I_{128})$ and produces a synthetic tabular row.
The architecture follows CTGAN~\cite{xu2019modeling}: a linear projection into three residual blocks (hidden width 256), then a final projection to $n_\text{feat}$ dimensions with per-column activations, sigmoid for
continuous columns, Gumbel Softmax~\cite{jang2016categorical} ($\tau{=}0.2$ during training, $\tau{=}10^{-5}$ at generation) for one-hot categorical groups, and sigmoid with a $0.5$ hard threshold for binary columns.
Each residual block consists of two fully connected layers with batch normalization applied on the main path: the first sub-layer follows the
order $\mathrm{FC} \to \mathrm{BN} \to \mathrm{ReLU}$, and the second applies
$\mathrm{FC} \to \mathrm{BN}$ without a trailing activation. The skip connection concatenates the block input $x$ directly to the transformed output $h$, so the output dimension grows as
$d_\text{out} = d_\text{hidden} + d_\text{in}$ per block; no normalization is applied to the skip path.

\paragraph{Student discriminator.}
$S_\phi$ adopts a TabTransGAN-style Transformer
architecture~\cite{zhang2025tabtransgan} in which every semantic column, continuous, one-hot categorical, and binary, is treated as an independent token.
Each token is produced by a dedicated linear projection of its column slice to a shared embedding space of dimension $d_\text{model}{=}64$; a learnable positional embedding is added per column.
The embedding dimension is chosen to be comfortably larger than the widest single-column one-hot group across our datasets while remaining computationally lightweight, a standard practice for tabular data with tens of features rather than hundreds.
The token sequence is processed by $L{=}3$ Transformer encoder layers with $H{=}4$ attention heads.
The head count follows directly from the embedding dimension: $d_\text{model}/H = 16$ dimensions per head, satisfying the divisibility constraint with a compact per-head space.
The depth $L{=}3$ is a deliberate, conservative choice that limits the discriminator's parametric complexity, a standard regularizing practice
for tabular domains and privacy-preserving frameworks alike, where over-parameterized models risk overfitting to the noisy private labels
produced by the teacher ensemble.
The encoder output is mean-pooled across the token dimension, passed through LayerNorm, and fed into a two-layer MLP head producing a scalar real/fake
logit.

\paragraph{Teacher ensemble.}
The teacher ensemble consists of $k$ Logistic Regression classifiers $\{T_i\}_{i=1}^k$, each trained on a distinct constituent partition $\mathcal{D}_i$ of the training data, where the partitions are disjoint and exhaustive:
\begin{equation*}
  \mathcal{D}_i \cap \mathcal{D}_j = \emptyset \quad \forall\, i \neq j,
  \qquad
  \bigcup_{i=1}^{k} \mathcal{D}_i = \mathcal{D}.
\end{equation*}
Each teacher $T_i$ is therefore exposed only to its own partition and has no knowledge of the records held by any other teacher, a property that enables the PATE privacy argument. Logistic Regression is chosen following the original PATE-GAN implementation~\cite{jordon2018pate}, and is well-suited to this setting due to its calibration stability and low sample complexity, both desirable when each partition contains only $|\mathcal{D}|/k$
training examples. At each outer iteration, $T_i$ is retrained on $\mathcal{D}_i$ (labeled real) versus fresh generator samples (labeled fake), keeping it calibrated to the current generator output.
 
% -------------------------------------------------
% 3.2 — Privacy Accounting
% --------------------------------------------------
\subsection{Privacy Accounting}
\label{sec:privacy}

\paragraph{Mechanism.}
We use the Gaussian Noisy Max (GNMax) mechanism~\cite{papernot2018scalable}:
Gaussian noise $\eta \sim \mathcal{N}(0, \sigma^2)$ is added to the
raw vote count $n_j$ before the majority label is released.
This replaces the Laplace mechanism of PATE-GAN, whose moments accountant produces numerical instabilities when the teacher vote margin is small.
The GNMax mechanism admits the RDP accountant
of~\cite{papernot2018scalable}, which is numerically stable across all vote margins.

\paragraph{Per-query RDP cost.}
The complementary error function is defined as
\begin{equation*}
  \mathrm{erfc}(x) = \frac{2}{\sqrt{\pi}} \int_{x}^{\infty} e^{-t^2}\,\mathrm{d}t,
\end{equation*}
which gives the probability that a standard Gaussian variable exceeds
$x\sqrt{2}$; it approaches $1$ as $x \to 0$ and decays to $0$ as
$x \to \infty$.
For each released label with vote tally $n_j$ out of $k$ teachers,
let $\mathrm{gap}_j = |n_j - k/2|$ and
\begin{equation*}
  q_j = \tfrac{1}{2}\,\mathrm{erfc}
        \!\left(\frac{\mathrm{gap}_j}{2\sigma}\right)
\end{equation*}
be the probability that Gaussian noise flips the teacher majority.
The data-dependent RDP cost at order $\alpha$ is given by~\cite{papernot2018scalable}:
\begin{equation}
  \varepsilon_\text{RDP}(\alpha;\,q_j,\sigma)
  = \frac{1}{\alpha-1}
    \log\!\left[
      (1-q_j)
      \!\left(\frac{1-q_j}{1 - q_j\,e^{2/\sigma^2}}\right)^{\!\alpha-1}
      + q_j\,e^{2(\alpha-1)/\sigma^2}
    \right]
  \label{eq:rdp}
\end{equation}
valid when $q_j\,e^{2/\sigma^2} < 1$; otherwise the accountant falls
back to the data-independent Gaussian RDP bound $\alpha/\sigma^2$, which is the
worst-case cost over all vote configurations.
When $q_j \approx 0$ (unanimous teachers, large gap), the
data-dependent cost approaches zero: a label released under strong
teacher consensus carries almost no privacy charge.

\paragraph{Composition and conversion.}
The accountant maintains a running sum of per-query RDP costs for each order $\alpha \in \{2, 3, \ldots, 511\}$:
\begin{equation}
  \varepsilon_\text{RDP,total}(\alpha)
  = \sum_{i=1}^{N_\text{released}}
    \varepsilon_\text{RDP}(\alpha;\,q_i,\sigma)
  \label{eq:composition}
\end{equation}
where $N_\text{released}$ is the total number of labels released to the student across all outer iterations.
This additive composition follows from the sequential composition property of RDP: if each query satisfies $(\alpha, \varepsilon_\text{RDP})$-RDP independently, their combination satisfies $(\alpha, \sum_i \varepsilon_\text{RDP,i})$-RDP.

After each outer iteration, the accumulated RDP cost is converted to a $(\varepsilon,\delta)$-DP guarantee using the conversion of Mironov~\cite{mironov2017renyi}:
\begin{equation}
  \hat\varepsilon(\delta)
  = \min_{\alpha \in [2,\,511]}
    \left[
      \varepsilon_\text{RDP,total}(\alpha)
      + \frac{\log(1/\delta)}{\alpha - 1}
    \right]
  \label{eq:conversion}
\end{equation}
The minimization over $\alpha$ selects the order that yields the tightest $(\varepsilon,\delta)$-DP bound for the given $\delta$; the $\log(1/\delta)/(\alpha-1)$ term is the conversion penalty that translates the RDP guarantee into the standard DP framework.
Training halts when $\hat\varepsilon \geq \varepsilon_\text{target}$; the privacy guarantee of the trained generator is the $(\hat\varepsilon, \delta)$ pair at that point.

\subsection{Training Algorithm}
\begin{algorithm}[H]
\caption{Train PATE-TabTransGAN}
\label{alg:training}
\begin{algorithmic}[1]

\Statex \textbf{Input:} Training data $\mathcal{D}$; number of teachers $k$;
  noise scale $\sigma$;
  batch size $B$; student steps per iteration $n_s$;
  privacy budget $(\varepsilon_{\mathrm{target}},\delta)$
\Statex \textbf{Output:} Generator parameters $\theta$;
  accumulated privacy cost $\hat\varepsilon$
\Statex \textbf{Notation:}
  $\mathcal{L}_{\mathrm{BCE}}(\hat{y},y)=-y\log\hat{y}-(1-y)\log(1-\hat{y})$;
  $\mathcal{A}(\sigma,k,\delta)$: GNMax RDP accountant where
  $\mathcal{A}.\textsc{RecordQuery}(\cdot)$ charges the privacy cost of one
  noisy aggregation and
  $\mathcal{A}.\textsc{GetEpsilon}()$ returns the current accumulated
  $\hat\varepsilon$

\Statex \textbf{procedure} \textsc{Train}
\State Partition $\mathcal{D}$ into $k$ disjoint shards
  $\mathcal{D}_1,\ldots,\mathcal{D}_k$
\State Initialise $G_\theta$, $S_\phi$;
  init accountant $\mathcal{A}(\sigma,k,\delta)$;\;
  $\hat\varepsilon\leftarrow 0$
\While{$\hat\varepsilon < \varepsilon_{\mathrm{target}}$}
  \For{$i=1$ \textbf{to} $k$}
    \State $\tilde{\mathcal{D}}_i\leftarrow\{G_\theta(z)\}^{|\mathcal{D}_i|}$,\;
      $z\sim\mathcal{N}(0,I)$
      \hfill$\triangleright$ generate fake shard
    \State Fit $T_i$ on $\mathcal{D}_i$ (label 1)
      $\cup\;\tilde{\mathcal{D}}_i$ (label 0)
  \EndFor
  \State $z^q\sim\mathcal{N}(0,I)^{B}$;\;
    $X^q\leftarrow G_\theta(z^q)$
    \hfill$\triangleright$ query batch
  \For{$j=1$ \textbf{to} $B$}
    \State $n_j\leftarrow\sum_{i=1}^{k}\mathbf{1}[T_i(x_j^q)=1]$
      \hfill$\triangleright$ votes for ``real''
    \State $Y_j\leftarrow\mathbf{1}[n_j+\mathcal{N}(0,\sigma^2)>k/2]$
      \hfill$\triangleright$ noisy label
  \EndFor
  \State $\mathcal{A}.\textsc{RecordQuery}(\{n_j\}_{j=1}^{B})$
    \hfill$\triangleright$ charge privacy cost
  \For{$s=1$ \textbf{to} $n_s$}
    \State $\phi\leftarrow\phi-
      \nabla_\phi\,\mathcal{L}_{\mathrm{BCE}}(S_\phi(X^q),\,Y)$
      \hfill$\triangleright$ student update
  \EndFor
  \State $\theta\leftarrow\theta-
    \nabla_\theta\,\mathcal{L}_{\mathrm{BCE}}(S_\phi(G_\theta(z)),\,\mathbf{1}_B)$
    \hfill$\triangleright$ generator update
  \State $\hat\varepsilon\leftarrow\mathcal{A}.\textsc{GetEpsilon}()$
    \hfill$\triangleright$ update accumulated cost
\EndWhile
\State \Return $G_\theta$,\;$\hat\varepsilon$

\end{algorithmic}
\end{algorithm}

\begin{figure}[t]
  \centering
  \includegraphics[width=0.82\textwidth]{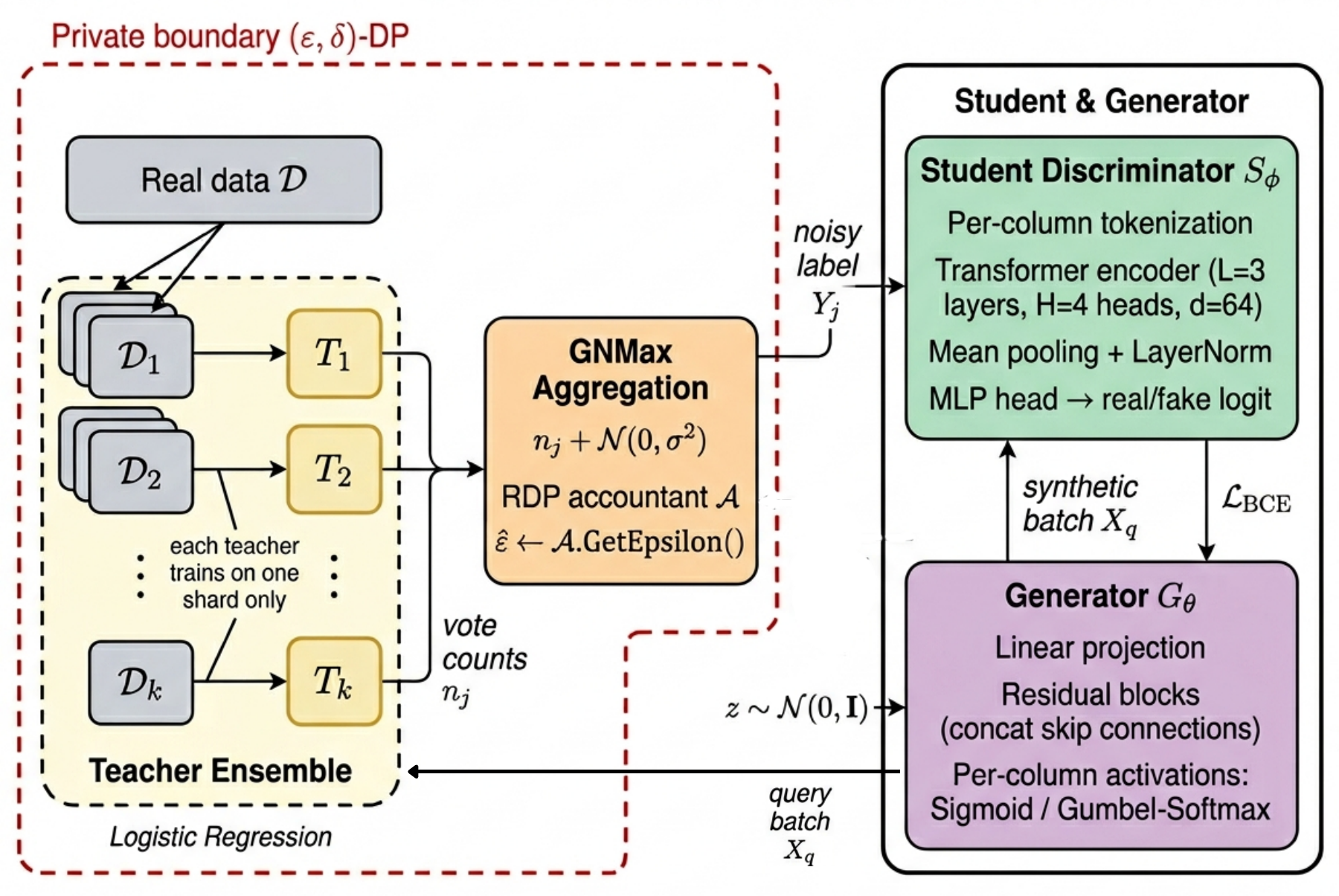}
  \caption{Overview of the PATE-TabTransGAN architecture. An ensemble of
           $k$ Logistic Regression teachers, each trained on a disjoint
           data shard, votes on generator-produced samples; votes are
           aggregated with Gaussian noise to produce private labels that supervise the Transformer-based student discriminator and the
           residual generator. The trained generator inherits a formal
           $(\varepsilon,\delta)$-DP guarantee by the post-processing
           property.}
  \label{fig:architecture}
\end{figure}

\section{Experiments}
\subsection{Goals}
\label{sec:goals}
The experiments are designed to address four research questions.
\emph{(i)} Does replacing the MLP student discriminator with a Transformer-based counterpart improve the quality of differentially private synthetic tabular data, as measured by downstream classification utility under matched privacy budgets? \emph{(ii)} Does PATE-TabTransGAN achieve higher utility than existing differentially private tabular generators (PATE-GAN,
DP-GAN, DP-CTGAN) under the same formal privacy constraints? \emph{(iii)} Is the best-performing downstream classifier consistent across datasets, and can its dominance be attributed to properties
of the privacy-preserving training pipeline? \emph{(iv)} Does the apparent AUCPR gap on the Adult dataset reflect a genuine synthesis deficit, or can it be explained by a difference in evaluation convention between pipelines?

\subsection{Experimental Setup}
\label{sec:setup}
\paragraph{Datasets.}
We evaluate on four publicly available tabular datasets with binary classification targets: Adult Income, Breast Cancer, Cardio, and Cervical.
Table~\ref{tab:datasets} summarizes the structure of each dataset; the minority-class prevalence is approximately $24\%$ (Adult), $30\%$ (Breast),
$50\%$ (Cardio), and $6\%$ (Cervical).
The baseline paper~\cite{fang2022dp} also evaluates on proprietary datasets that are not publicly accessible, which we exclude from our comparison.

\begin{table}[H]
\centering
\caption{Dataset structure.}
\label{tab:datasets}
\begin{tabular}{lccrr}
\toprule
Dataset & Objects & Attributes & Class 0 & Class 1 \\
\midrule
Adult           & 48{,}842 & 15 & ${\leq}50\text{K}$: 37{,}155 & ${>}50\text{K}$: 11{,}687 \\
Cardio  & 70{,}000 & 13 & Absent: 35{,}021             & Present: 34{,}979         \\
Cervical &    858   & 36 & Negative: 803                & Positive: 55              \\
Breast   &    286   & 10 & No Recurrence: 218           & Recurrence: 68            \\
\bottomrule
\end{tabular}
\end{table}
\paragraph{Baselines.}
We compare against PATE-GAN~\cite{jordon2018pate},
DP-GAN~\cite{xie2018differentially}, and DP-CTGAN~\cite{fang2022dp},
with results taken directly from~\cite{fang2022dp}.

\paragraph{Privacy configuration.}
All experiments use $(\varepsilon, \delta)$-DP with $\delta = 10^{-5}$
and per-dataset privacy budgets matching~\cite{fang2022dp},
summarised in Table~\ref{tab:privacy}.

\begin{table}[H]
\centering
\caption{Per-dataset privacy budgets ($\delta = 10^{-5}$ throughout).}
\label{tab:privacy}
\begin{tabular}{lcccc}
\toprule
         & Adult & Breast & Cardio & Cervical \\
\midrule
$\varepsilon$ & 3 & 4 & 2 & 2 \\
\bottomrule
\end{tabular}
\end{table}

\paragraph{Evaluation protocol.}
We follow the TSTR protocol of~\cite{fang2022dp}: five classifiers (Logistic Regression, Decision Tree, Random Forest, AdaBoost, MLP) are trained on synthetic data and evaluated on real held-out test sets, reporting AUROC and AUCPR averaged over five independent runs.

\paragraph{Implementation. } The implementation of methods and the experimental environment are done using the Python programming language. Complete source code, sufficient to repeat the experiments, was made available online \footnote{\url{https://anonymous.4open.science/r/PATE-TabTransGAN-B467/README.md}}.

\subsection{Comparative Utility Results}
\label{sec:comparative}
All PATE-TabTransGAN entries in Tables~\ref{tab:auroc} and~\ref{tab:aucpr} are reported as the mean over five independent training runs at the per-dataset privacy budgets specified in Section~\ref{sec:setup}.
Baseline values for PATE-GAN, DP-GAN, and DP-CTGAN are reproduced from~\cite{fang2022dp}; the dash in the PATE-GAN row of the Breast column reflects that this combination is not reported in the original study.
\begin{table}[H]
\centering
\caption{Mean AUROC across five downstream classifiers (TSTR).
         Best result per dataset in \textbf{bold}.}
\label{tab:auroc}
\begin{tabular}{lcccc}
\toprule
Dataset  & DP-GAN & PATE-GAN & DP-CTGAN        & PATE-TabTransGAN \\
\midrule
Adult    & 0.5189 & 0.5008   & 0.5097          & $\mathbf{0.6487}$ \\
Breast   & 0.3740 & --       & 0.6601          & $\mathbf{0.7454}$ \\
Cardio   & 0.4876 & 0.4648   & $\mathbf{0.6827}$ & $0.6793$ \\
Cervical & 0.4617 & 0.4920   & 0.5084          & $\mathbf{0.5451}$ \\
\bottomrule
\end{tabular}
\end{table}
\paragraph{AUROC results.}
Table~\ref{tab:auroc} shows that PATE-TabTransGAN obtains the highest mean AUROC on three of the four datasets: Adult ($0.6487$), Breast ($0.7454$),
and Cervical ($0.5451$), with margins of $0.130$, $0.085$, and $0.037$ over the strongest reproduced baseline on each dataset.
On Cardio, our model attains $0.6793$, marginally below DP-CTGAN's reported value of $0.6827$; the difference of $0.0034$ is small relative to typical
run-to-run variation, and we therefore regard the two methods as effectively matched on this benchmark.
The earlier baselines (DP-GAN and PATE-GAN) cluster near $0.5$ AUROC on Adult, Cardio, and Cervical, indicating near-random ranking under the matched
privacy budgets; DP-CTGAN improves on this floor only on Cardio.
PATE-TabTransGAN, by contrast, exceeds $0.54$ AUROC across all benchmarks. Comparing directly against PATE-GAN, which shares the same teacher-student
training structure but uses an MLP student, PATE-TabTransGAN improves AUROC by $0.148$ on Adult, $0.081$ on Cardio, and $0.053$ on Cervical, providing direct evidence that the Transformer student captures inter-column structure
that the MLP student does not.
\begin{table}[h]
\centering
\caption{Mean AUCPR across five downstream classifiers.
         Best result per dataset in \textbf{bold}.}
\label{tab:aucpr}
\begin{tabular}{lcccc}
\toprule
Dataset  & DP-GAN & PATE-GAN & DP-CTGAN        & PATE-TabTransGAN \\
\midrule
Adult    & 0.7633 & 0.7500   & $\mathbf{0.7653}$ & $0.3959$ \\
Breast   & 0.5520 & --       & $\mathbf{0.7377}$ & $0.5953$ \\
Cardio   & 0.4987 & 0.4825   & $\mathbf{0.6709}$ & $0.6707$ \\
Cervical & 0.0776 & 0.1325   & $0.1343$          & $\mathbf{0.1415}$ \\
\bottomrule
\end{tabular}
\end{table}
\paragraph{AUCPR results.}
The picture under AUCPR (Table~\ref{tab:aucpr}) is more heterogeneous.
PATE-TabTransGAN matches DP-CTGAN on Cardio to within $0.0002$ ($0.6707$ versus $0.6709$) and obtains the highest AUCPR on Cervical ($0.1415$); we note, however, that all methods score low in absolute terms on this strongly imbalanced benchmark, so the Cervical ranking
should be read in that context. On Breast, our model trails DP-CTGAN by $0.142$ ($0.5953$ versus
$0.7377$) while remaining above DP-GAN.
The most significant deviation is observed in the Adult dataset, where
PATE-TabTransGAN yields a result of $0.3959$ compared to the reported reference values, which range from $0.75$ to $0.77$.
This combination is internally unusual: on the same dataset, PATE-TabTransGAN leads all baselines on AUROC by a margin of $0.13$, and a method that ranks examples substantially better than its competitors does not
typically produce AUCPR more than $0.35$ below them on the same task.
Rather than presenting the Adult AUCPR figure as a direct measurement of synthesis quality, we treat it as a result requiring separate empirical examination; the source of the discrepancy is investigated in Section~\ref{sec:labelswap}.

%---------------------------------------------------
% 4.4 — Sensitivity of AUCPR to Positive-Class Convention
%---------------------------------------------------
\subsection{Sensitivity of AUCPR to Positive-Class Convention}
\label{sec:labelswap}

AUCPR is bounded below by the positive-class prevalence under random ranking: a chance-level classifier attains expected AUCPR equal to the fraction of positive examples in the evaluation set.
This property means that a mismatch in positive-class convention between two evaluation pipelines produces a large, systematic gap in reported AUCPR that is entirely unrelated to synthesis quality.
The Adult dataset illustrates this sensitivity sharply.

\paragraph{Prevalence argument.}
Our pipeline treats the income ${>}$\$50K class (prevalence $\approx 24\%$) as positive, following standard practice for imbalanced classification.
A chance-level ranker then attains expected AUCPR $\approx 0.24$; with the majority class (${\leq}$\$50K, prevalence $\approx 76\%$) as positive, it attains $\approx 0.76$.
The reported baseline AUCPR of $0.75$--$0.77$ is internally consistent only if positive prevalence is near $0.76$, suggesting that~\cite{fang2022dp} treats the majority class as positive.
Our own AUROC advantage of $0.13$ on Adult makes a genuine synthesis deficit an unlikely explanation for the $0.35$ AUCPR gap; a convention difference is a more parsimonious account.

\begin{figure}[H]
  \centering
  \includegraphics[width=0.82\textwidth]{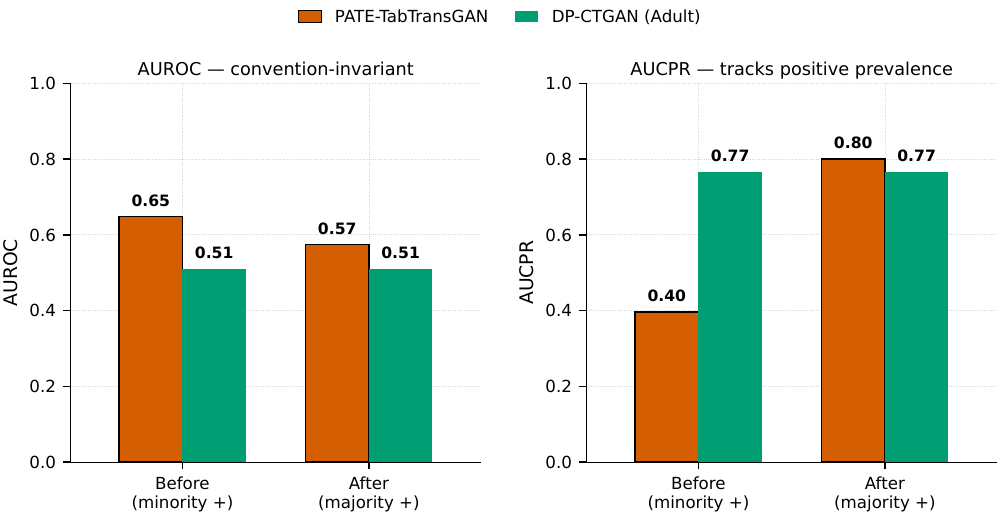}
  \caption{Effect of positive-class convention on Adult AUROC and AUCPR at
           $\varepsilon = 3$.
           \emph{Before}: minority class (income ${>}$\$50K,
           $\approx 24\%$) as positive (our default).
           \emph{After}: majority class as positive, matching the convention
           implied by~\cite{fang2022dp}.
           Reproduced DP-CTGAN values shown in both panels.}
  \label{fig:labelswap}
\end{figure}

\paragraph{Controlled re-evaluation.}
We re-ran the Adult pipeline with the majority class as positive at $\varepsilon = 3$, $\delta = 10^{-5}$, all other hyperparameters fixed; downstream classifiers were re-fit under the new target.
Figure~\ref{fig:labelswap} reports the outcome with reproduced DP-CTGAN values for reference.
PATE-TabTransGAN's AUCPR shifts from $0.40$ to $0.80$, within the baseline range $[0.75, 0.77]$.
Its AUROC moves from $0.65$ to $0.57$: AUROC is invariant under label inversion for a fixed score function, so this $0.08$ change reflects classifier retraining variance rather than a metric effect, and our model remains above DP-CTGAN's $0.51$ in both panels.

\paragraph{Scope and limitations.}
The re-evaluation shows the conventional difference is sufficient to explain the observed AUCPR gap; this does not establish that it is the cause.
The baseline preprocessing pipeline is not directly inspectable, and alternative explanations, such as different feature preprocessing, class
weighting, or threshold calibration, cannot be ruled out.
The argument applies only to Adult and all other AUCPR comparisons in Table~\ref{tab:aucpr} stand as reported.
%We retain the minority-positive convention in our headline results. We recommend that future work on imbalanced tabular benchmarks explicitly report which class is treated as positive, so that AUCPR comparisons across pipelines remain unambiguous.
In our main findings, we maintain the convention of treating the minority group as the reference group.
However, comparative studies should clearly indicate which class is treated as the positive class, so that comparisons of prediction quality metrics (such as AUCPR, as analyzed in this paper) are unambiguous.
% --------------------------------------------------
% 4.6 — Downstream Classifier Analysis
% --------------------------------------------------
\subsection{Downstream Classifier Analysis}
\label{sec:classifier}
Logistic Regression attains the highest mean score on both metrics. On AUROC, it is best or tied-best across all datasets (tied with three classifiers at $0.74$ on Breast, tied with Random Forest at $0.59$ on Cervical); on AUCPR, it leads on three datasets and trails Random Forest by $0.01$ on Breast. Decision Tree is uniformly weakest (mean AUROC $0.58$, AUCPR $0.36$); AdaBoost, MLP, and Random Forest fall between, with Random Forest second on both means.
Two factors plausibly drive this ordering.
First, the student is supervised through noisy aggregation of Logistic Regression teachers, and the generator is, in turn, optimized against that student; this propagates a linear decision signal through the training loop, so the surviving synthetic distribution preserves linearly separable structure with respect to the target more faithfully than higher-order structure, which a downstream linear classifier recovers most directly.
Second, synthetic columns carry residual artifacts (quantization, smoothed tails, and ordinal mass redistribution) that higher-variance learners are more inclined to fit.
Decision Tree, an unregularized axis-aligned partitioner, memorizes such artifacts and fails to transfer to the real test distribution; Random Forest's bootstrap averaging partially neutralizes this failure mode, accounting for its second-place position.

\begin{figure}[t]
  \centering
  \includegraphics[width=\textwidth]{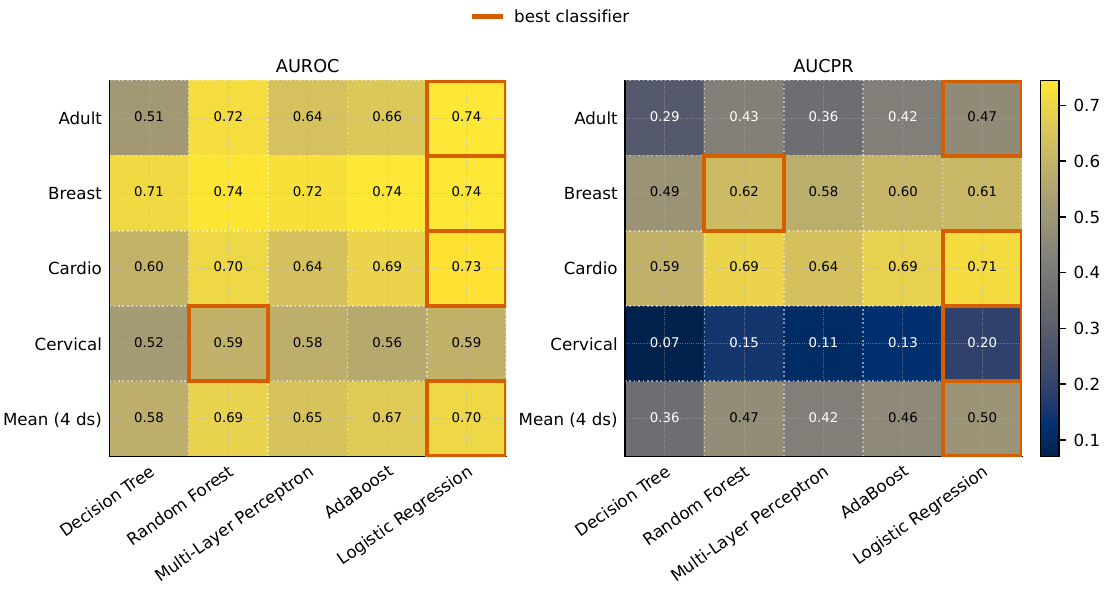}
  \caption{Per-classifier AUROC (left) and AUCPR (right) for
           PATE-TabTransGAN across the four datasets, mean over
           five runs. Outlined cells mark the best classifier per
           row.}
  \label{fig:classifiers}
\end{figure}

%---------------------------------------------------
% 4.6 — Discussion
%---------------------------------------------------
\subsection{Discussion}
\label{sec:discussion}

\paragraph{Answers to the research questions.}
Returning to the four questions of Section~\ref{sec:goals}:
\emph{(i)}~Yes, the Transformer student improves AUROC over the MLP-based PATE-GAN by $0.053$--$0.148$ across Adult, Cardio, and
Cervical, isolating the architectural change as the cause.
\emph{(ii)}~Yes on ranking, PATE-TabTransGAN attains the best or tied-best AUROC on all four datasets, while earlier baselines sit near $0.5$; on AUCPR it matches DP-CTGAN on Cardio, leads on Cervical, and trails on Breast.
\emph{(iii)}~Yes, Logistic Regression is consistently best or tied-best, a consequence of the Logistic Regression teacher ensemble propagating a linear decision signal through the training loop.
\emph{(iv)}~The apparent Adult AUCPR gap is consistent with a difference in positive-class convention rather than a genuine synthesis deficit,
re-evaluating under the majority-positive convention raises our AUCPR from $0.40$ to $0.80$, which is sufficient to account for the gap,
though preprocessing differences in the baseline pipeline cannot be ruled out see Section~\ref{sec:labelswap}.

\paragraph{Utility under matched privacy.}
While the privacy-utility trade-off remains a fundamental challenge for private data synthesis, PATE-TabTransGAN significantly narrows this gap.
Under matched privacy budgets, it attains the highest mean AUROC on three of four benchmarks and is statistically indistinguishable from DP-CTGAN on
Cardio; AUCPR is competitive on Cardio and Cervical, and lower on Breast, with the Adult gap addressed in Section~\ref{sec:labelswap}.
This is consistent with the architectural hypothesis: a Transformer student carries inter-column modeling capacity into PATE without weakening its formal guarantee, and that capacity translates more directly into ranking quality than into precision--recall.

\paragraph{Classifier dependence and inductive bias.}
The dominance of Logistic Regression can be traced to the inductive bias of the PATE pipeline: the student discriminator is supervised through noisy
aggregation of Logistic Regression teachers, and the generator is in turn optimised against that student, propagating a linear decision signal through the training loop.
The surviving synthetic distribution therefore preserves linearly separable structure with respect to the target more faithfully than higher-order
structure, which a downstream linear classifier recovers most directly.
Higher-capacity learners, conversely, tend to overfit residual artifacts in the synthetic data (quantisation, smoothed tails, and ordinal mass
redistribution).
Classifier-dependent performance is therefore not purely a property of the downstream models, but partly reflects an inductive bias imposed by the
privacy-preserving training pipeline, suggesting fidelity directions beyond marginal alignment.

\paragraph{Limitations.}
Three limitations bound our claims.
\emph{(i)} The Adult label-convention analysis relies on a controlled
re-evaluation in our pipeline; the baseline pipeline is not directly
inspectable, and preprocessing differences cannot be excluded.
\emph{(ii)} All experiments adopt the per-dataset privacy budgets and evaluation protocol of~\cite{fang2022dp} to ensure a like-for-like
comparison; this configuration is not necessarily the most favorable to our method, and a setting tuned specifically to PATE-TabTransGAN may yield
further improvements.
\emph{(iii)} Evaluation covers four small-to-medium binary-target datasets; behaviour on larger, higher-arity, or multi-class settings remains open.

%---------------------------------------------------
% 5 — Conclusion
%---------------------------------------------------
\section{Conclusion}
\label{sec:conclusion}

We presented PATE-TabTransGAN, a differentially private tabular generator
that replaces the MLP student discriminator of PATE-GAN with a Transformer-based counterpart, explicitly modeling inter-column dependencies
while preserving the formal $(\varepsilon,\delta)$-DP guarantee via the post-processing property.
A GNMax RDP accountant is adopted in place of the original Laplace moments accountant to ensure numerically stable privacy accounting across all teacher vote margins.
Under matched $(\varepsilon,\delta)$ budgets on four benchmarks, the method attains the highest mean AUROC on Adult, Breast, and Cervical, and is
statistically indistinguishable from DP-CTGAN on Cardio.
On AUCPR, the method matches DP-CTGAN on Cardio, leads on Cervical, and trails on Breast; the apparent gap on Adult is shown by a controlled
re-evaluation to be consistent with a difference in positive-class convention rather than a synthesis deficit, and we recommend that future imbalanced
tabular benchmarks explicitly report which class is treated as positive.
The consistent dominance of Logistic Regression among downstream classifiers suggests that the privacy-preserving training pipeline may impose an
inductive bias on the synthetic distribution, which we identify as a direction for further investigation.

\vspace{0.5cm}

\noindent\paragraph{\textbf{Declaration of using AI tools. }}We acknowledge the use of Grammarly and Claude to polish the selected parts of the manuscript.

%\begin{credits}

%\subsubsection{\ackname}

%\subsubsection{\discintname}

%\end{credits}

\bibliographystyle{splncs04}
\bibliography{references}

\begin{thebibliography}{10}
\providecommand{\url}[1]{\texttt{#1}}
\providecommand{\urlprefix}{URL }
\providecommand{\doi}[1]{https://doi.org/#1}

\bibitem{abadi2016deep}
Abadi, M., Chu, A., Goodfellow, I., McMahan, H.B., Mironov, I., Talwar, K., Zhang, L.: Deep learning with differential privacy. In: Proceedings of the 2016 ACM SIGSAC Conference on Computer and Communications Security. pp. 308--318 (2016)

\bibitem{angiulli2007fast}
Angiulli, F.: Fast nearest neighbor condensation for large data sets classification. IEEE Transactions on Knowledge and Data Engineering  \textbf{19}(11),  1450--1464 (2007)

\bibitem{carlini2022quantifying}
Carlini, N., Ippolito, D., Jagielski, M., Lee, K., Tramer, F., Zhang, C.: Quantifying memorization across neural language models. In: The Eleventh International Conference on Learning Representations (2022)

\bibitem{carlini2021extracting}
Carlini, N., Tramer, F., Wallace, E., Jagielski, M., Herbert-Voss, A., Lee, K., Roberts, A., Brown, T., Song, D., Erlingsson, U., et~al.: Extracting training data from large language models. In: 30th USENIX security symposium (USENIX Security 21). pp. 2633--2650 (2021)

\bibitem{dwork2011firm}
Dwork, C.: A firm foundation for private data analysis. Communications of the ACM  \textbf{54}(1),  86--95 (2011)

\bibitem{dwork2006calibrating}
Dwork, C., McSherry, F., Nissim, K., Smith, A.: Calibrating noise to sensitivity in private data analysis. In: Theory of Cryptography Conference. pp. 265--284. Springer (2006)

\bibitem{dwork2014algorithmic}
Dwork, C., Roth, A.: The algorithmic foundations of differential privacy. Foundations and trends{\textregistered} in theoretical computer science  \textbf{9}(3-4),  211--487 (2014)

\bibitem{fang2022dp}
Fang, M.L., Dhami, D.S., Kersting, K.: Dp-ctgan: Differentially private medical data generation using ctgans. In: International conference on artificial intelligence in medicine. pp. 178--188. Springer (2022)

\bibitem{feldman2020does}
Feldman, V.: Does learning require memorization? a short tale about a long tail. In: Proceedings of the 52nd annual ACM SIGACT symposium on theory of computing. pp. 954--959 (2020)

\bibitem{fredrikson2015model}
Fredrikson, M., Jha, S., Ristenpart, T.: Model inversion attacks that exploit confidence information and basic countermeasures. In: Proceedings of the 22nd ACM SIGSAC Conference on Computer and Communications Security. pp. 1322--1333 (2015)

\bibitem{gaboardi2014dual}
Gaboardi, M., Arias, E.J.G., Hsu, J., Roth, A., Wu, Z.S.: Dual query: Practical private query release for high dimensional data. In: International Conference on Machine Learning. pp. 1170--1178. PMLR (2014)

\bibitem{gulati2023tabmt}
Gulati, M., Roysdon, P.: Tabmt: Generating tabular data with masked transformers. Advances in Neural Information Processing Systems  \textbf{36},  46245--46254 (2023)

\bibitem{jang2016categorical}
Jang, E., Gu, S., Poole, B.: Categorical reparameterization with gumbel-softmax. arXiv preprint arXiv:1611.01144  (2016)

\bibitem{jordon2018pate}
Jordon, J., Yoon, J., Van Der~Schaar, M.: Pate-gan: Generating synthetic data with differential privacy guarantees. In: International Conference on Learning Representations (2018)

\bibitem{mironov2017renyi}
Mironov, I.: R{\'e}nyi differential privacy. In: 2017 IEEE 30th computer security foundations symposium (CSF). pp. 263--275. IEEE (2017)

\bibitem{papernot2016semi}
Papernot, N., Abadi, M., Erlingsson, U., Goodfellow, I., Talwar, K.: Semi-supervised knowledge transfer for deep learning from private training data. arXiv preprint arXiv:1610.05755  (2016)

\bibitem{papernot2018scalable}
Papernot, N., Song, S., Mironov, I., Raghunathan, A., Talwar, K., Erlingsson, {\'U}.: Scalable private learning with pate. arXiv preprint arXiv:1802.08908  (2018)

\bibitem{xie2018differentially}
Xie, L., Lin, K., Wang, S., Wang, F., Zhou, J.: Differentially private generative adversarial network. arXiv preprint arXiv:1802.06739  (2018)

\bibitem{xu2019modeling}
Xu, L., Skoularidou, M., Cuesta-Infante, A., Veeramachaneni, K.: Modeling tabular data using conditional gan. Advances in neural information processing systems  \textbf{32} (2019)

\bibitem{zhang2025tabtransgan}
Zhang, H., Jing, Y., Zhang, F., Li, Z., Wang, X.S., Chen, Z., Lv, C.: Tabtransgan: A hybrid approach integrating gan and transformer architectures for tabular data synthesis. Information Processing \& Management  \textbf{62}(5),  104220 (2025)

\bibitem{zhang2021privsyn}
Zhang, Z., Wang, T., Li, N., Honorio, J., Backes, M., He, S., Chen, J., Zhang, Y.: Privsyn: Differentially private data synthesis. In: 30th USENIX Security Symposium (USENIX Security 21). pp. 929--946 (2021)

\bibitem{zhao2023dataset}
Zhao, B., Bilen, H.: Dataset condensation with distribution matching. In: Proceedings of the IEEE/CVF winter conference on applications of computer vision. pp. 6514--6523 (2023)

\end{thebibliography}

\end{document}